\def\year{2020}\relax
\documentclass[letterpaper]{article} 
\usepackage{aaai20}  
\usepackage{times}  
\usepackage{helvet} 
\usepackage{courier}  
\usepackage[hyphens]{url}  
\usepackage{graphicx} 
\usepackage{graphicx}  
\usepackage{booktabs}
\usepackage{mathtools}

\usepackage{amsmath}
\usepackage{amssymb}
\usepackage{bm}
\usepackage{enumitem}
\usepackage{algorithmic,algorithm}  

\usepackage[usenames, dvipsnames, table]{xcolor}

\definecolor{col2}{RGB}{224,56,189}
\definecolor{col3}{RGB}{0, 211, 234}
\definecolor{col1}{RGB}{234,180,160}

\title{Differentiable Grammars for Videos}
\author{AJ Piergiovanni, Anelia Angelova, Michael S. Ryoo\\
Robotics at Google\\
\texttt{\{ajpiergi,anelia,mryoo\}@google.com}}
 
\begin{document}

\maketitle

\begin{abstract}
This paper proposes a novel algorithm which learns a formal regular grammar from real-world continuous data, such as videos.
Learning latent terminals, non-terminals, and production rules directly from continuous data allows the construction of a generative model capturing sequential structures with multiple possibilities.
Our model is fully differentiable, and provides easily interpretable results which are important in order to understand the learned structures.
It outperforms the state-of-the-art on several challenging datasets and is more accurate for forecasting future activities in videos. We plan to open-source the code.\footnote{https://sites.google.com/view/differentiable-grammars}

\end{abstract}

\noindent Learning a formal grammar from continuous, unstructured data is a challenging problem. This is especially challenging when the elements (i.e., terminals) of the grammar to be learned are not symbolic or discrete~\cite{chomsky1956three,chomsky1959on}, but are higher dimensional vectors, such as representations from 
real world data sequences such as videos.

Simultaneously, addressing such challenges is necessary for better automated understanding of sequential data. In video understanding, such as activity detection, a convolutional neural network (CNN) (e.g., \cite{carreira2017quo}) generates a representation abstracting local spatio-temporal information at every time step, forming a temporal sequence of representations.
Learning a grammar reflecting sequential changes in video representations will enable explicit and high-level modeling of temporal structure and relationships between multiple occurring events in videos. 
\begin{figure}
    \centering
    \includegraphics[width=\linewidth]{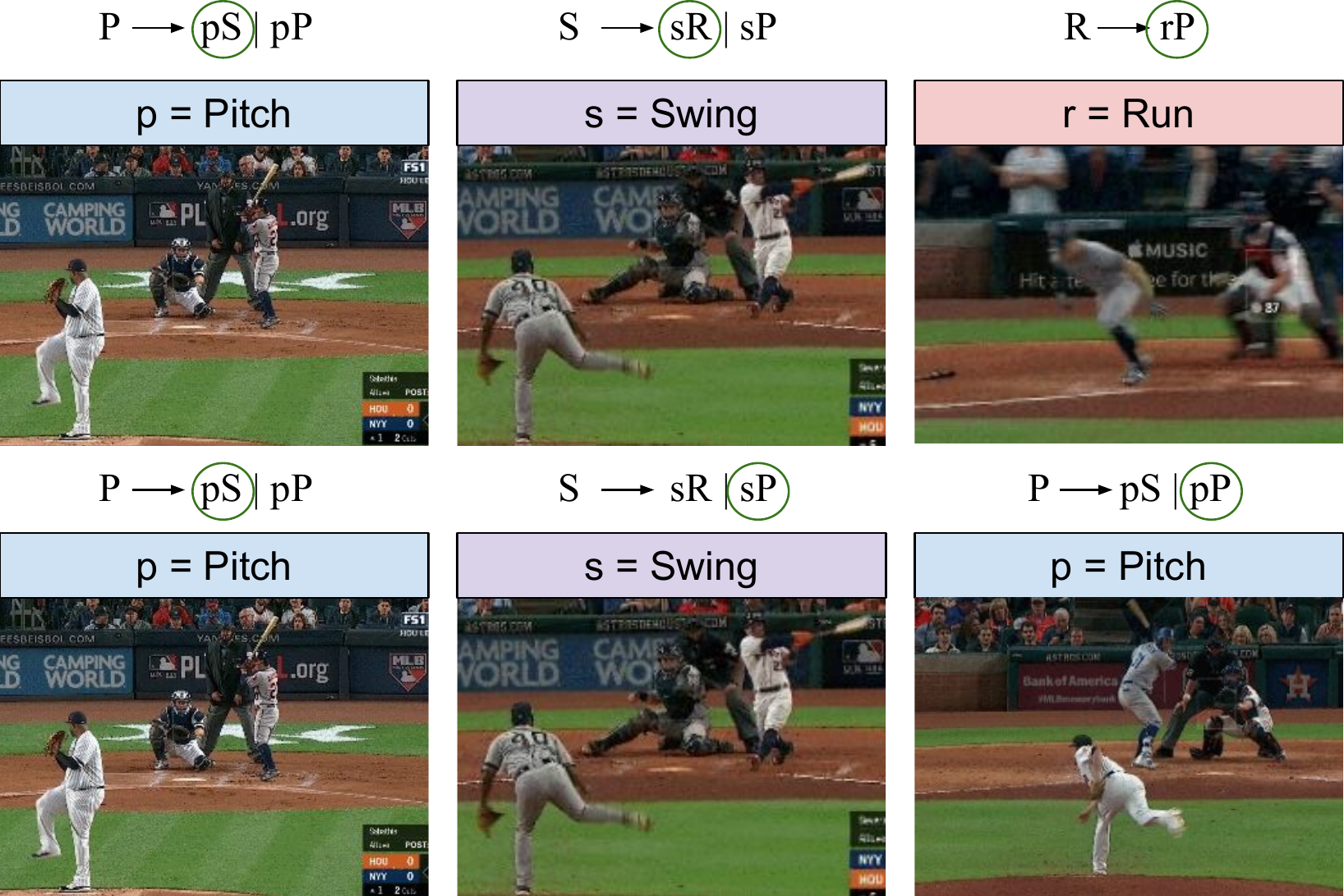}
    \caption{Example regular grammar giving the sequence of possible activities in a baseball video. For example, a swing (s) only occurs after a pitch (p). After a swing (s), both pitch (p) and run (r) are possible continuations of the activity.}
    \label{fig:baseball-example}
\end{figure}

In this paper, we propose a new approach of modeling a formal grammar for videos in terms of learnable and differentiable neural network functions. The objective is to formulate not only the terminals and non-terminals of our grammar as learnable representations but also the production rules, which are generated here as differentiable functions.
We provide the loss function to train our differentiable grammar directly from data, and present methodologies to take advantage of it for recognizing and forecasting sequences\footnote{Technically the grammar we are learning is a {\it stochastic} regular grammar, we use regular grammar for simplicity in the text.}.
Rather than focusing on non-terminals and production rules to generate or parse symbolic data (e.g., text strings), our approach allows learning of grammar representations directly on top of higher-dimensional data stream (e.g., representation vector sequences). We confirm such capability experimentally by focusing on learning a differentiable regular grammar from continuous representations, which can be applied to any sequential data including outputs of 3-D CNNs.

Fig.~\ref{fig:baseball-example} shows an example of consecutive events in video which are learned as grammar rules, e.g. "pitch" is followed by a "swing", which in turn can develop in a "run" event or a another "pitch".
This not only allows for better recognition of human activities from videos by enforcing the learned grammar to local-level detections, but also enables forecasting of future representations based on the learned production rules.
It also provides semantic interpretability of the video recognition and prediction process.

Our differentiable grammar could be interpreted as a particular form of recurrent neural network (RNN). The main difference to the standard RNNs such as LSTMs and GRUs \cite{hochreiter1997,cho2014learning} is that our grammar explicitly maintains a set of non-terminal representations, in contrast to having a single hidden representation in standard RNNs, and learns multiple distinct production rules per non-terminal. This not only makes the learned model more semantically interpretable, but also allows learning of temporal structures with multiple sequence possibilities. Our grammar, learned with a randomized production rule selection function, considers multiple transitions between abstract non-terminals when matching it with the input sequences as well as when generating multiple possible future sequences.

In the experiments, we observe much better performance on state-of-the-art on detection tasks on three video datasets, and much better accuracy on video forecasting.
We also experimentally compare our grammar with previous related models including LSTMs~\cite{hochreiter1997} and Neural Turing Machines (NTMs)~\cite{graves2014neural} in the experiments section.


The primary contributions of our work are:

\begin{itemize}
  \item We propose a \textbf{fully differentiable} model that is able to learn the structure (terminals, non-terminals, and production rules) of a regular grammar. This is done while still maintaining the \textbf{interpretability} of the learned model and its representations.
  \item We show that the model is able to achieve better results on \textbf{forecasting} of human activities which are to occur subsequently in videos.
  \item We confirm that the approach works on sequential real-world datasets, and outperforms the state-of-the-art on \textbf{challenging benchmarks}.
  
\end{itemize}

The goal of this work is to provide to the research community a neural differentiable 
grammar-based matching and prediction for video analysis, which is also applicable to other domains. We observe that learning a grammar for continuous video data greatly benefits its automated understanding. The results are interpretable which is very important for real-life decision making scenarios. Furthermore, it can predict future events with higher accuracy, which is crucial for anticipation and reaction to future actions, for example for an autonomous robot which interacts with humans in dynamic environments.

\section{Related work}

Chomsky grammars~\cite{chomsky1956three,chomsky1959on} are designed to represent functional linguistic relationships. They have found wide applications in defining programming languages, natural language understanding, and understanding of images and videos \cite{socher2011parsing}. In its original form, a grammar is composed of production rules generating discrete symbols. Machine learning of formal grammar given training data has been traditionally known as grammar induction \cite{fu1997}. Here, our main motivation is to newly design a differentiable version of a grammar representation whose parameters could be optimized with a standard backpropagation (together with the other components) for its learning.


There are early works exploring extracting grammars/state machines from trained RNNs \cite{kolen1994fool,boden2000context,tivno1998finite}. Other works have attempted to learn `neural push down-automata' to learn context-free grammars \cite{sun2017neural} or neural Turing Machines \cite{graves2014neural}. However, these works only explored simple toy experiments, and were not tested on real-world data. Our work is also related to the fields of differentiable rule learning~\cite{yang2017differentiable,evans2018learning}, the literature review of which goes beyond the scope of the paper.

Some works have explored learning more explicit structures by forcing states to be discrete and uses pseudo-gradients to learn grammatical structures \cite{zeng1994discrete}. However, they still rely on a standard RNN to learn model the sequences. It has also been found that LSTMs/RNNs are able to learn grammars \cite{gers2001lstm,giles1995learning,das1992learning}. Unlike these works, we design a neural network architecture that is able to explicitly model the structure of a grammar, which leads to much easier interpretability.

Other works have explored using neural networks to learn a parser. \cite{socher2011parsing} parse scenes by learning to merge representations. \cite{mayberry1999sardsrn} learn a shift-reduce neural network parser and \cite{chen2014fast} learn a dependency parser as a neural network. While these works learn grammar structures, they are generally difficult to interpret. 

Within the activity recognition domain, regular and context-free grammars have been tranditionally used to parse and understand videos \cite{moore2002recognizing,pirsiavash2014parsing,ivanov2000recognition,ryoo2009semantic,si2011unsupervised}. Other works have extended CFGs such as attribute grammars \cite{joo2006attribute} or using context-sensitive constraints and interval logic \cite{brendel2011probabilistic,kwak2014line}. We similarly develop our approach for video understanding tasks. However, none of such grammars were differntiable, preventing their end-to-end learning.

\section{Background}

A formal grammar $G$ is defined with four elements: $G = (V, \Sigma, P, S)$
where $V$ is a finite set of non-terminals, $\Sigma$ is a finite set of terminals, $P$ is a finite set of production rules, and $S$ is the starting non-terminal.

In a regular grammar, the production rules $P$ are in the following forms:
\begin{equation}
\begin{split}
A \hspace{0.07cm} &\to aB  \\ 
A \hspace{0.07cm} &\to a  \\
A \hspace{0.07cm} &\to \epsilon 
\end{split}
\end{equation}
where $A$ and $B$ are non-terminals in $V$, $a$ is any terminal in $\Sigma$, and $\epsilon$ denotes an empty string. A regular grammar is a type 3 formal grammar in the Chomsky hierarchy.

In this paper, we follow this traditional regular grammar definition, while extending it by making its terminals, non-terminals, and production rules represented in terms of differentiable neural network functions. 



\section{Approach}

\subsection{Formulation}

We model our formal grammar in terms of latent representations and differentiable functions mapping to representations. The parameters of our functions define production rules, which are learned together with the terminal and non-terminal representations.
For example, in the context of videos terminals can be per-frame activity labels of a video (multi-labels are allowed too), whereas the role of the production grammar rules is to learn plausible (stochastic) transitions between activities in time (Fig.~\ref{fig:baseball-example}). The non-terminals are learned internally within the grammar and do not have specific interpretation.

Each non-terminal in $V$ is a latent representation with fixed dimensionality, whose actual values are learned based on the training data. Each terminal in $\Sigma$ corresponds to a video representation that could be obtained at every time step, such as a vector with activity class predictions. This has to be learned as well. Our production rules are represented as a pair of two functions:
\begin{itemize}
\item $f$: a function that maps each non-terminal in $V$ (e.g., $A$) to a subset of production rules (i.e., the rules that `expand' the current non-terminal) $\{p_i\} \subset P$.
\item $g$: a function that maps each rule $p_i$ to a terminal (e.g., $a$) and the next non-terminal (e.g., $B$). 
\end{itemize}
\begin{equation}
\begin{split}
f:& V \rightarrow \{P \} \\
g:& P \rightarrow (V, \Sigma).
\end{split}
\end{equation}

The combination of the two functions effectively captures multiple production rules per non-terminal, such as ``$A \rightarrow aB$'' and ``$A \rightarrow aA$''. The starting non-terminal $S$ is learned to be one of the latent representations in $V$. The functions are learned from data.


These form a straight forward (recursive) generative model, which starts from the starting non-terminal $S = v^0$ and iteratively generates a terminal at every time step. 
The system internally keeps states of possible rules and non-terminals and learns their relationships with terminals according to data, starting from a feature representation of the video, non-terminal `states' $v$ are learned, each one of which learns to generate a number of rules. Each rule is then expanded into a set of non-terminal and terminal, where the terminal can now be compared to the actually observed data (terminal value). This process continues recursively. In the appendix, we illustrate this process and how the loss is computed.

Representing our production rules as functions allows us to model the generation of a sequence (i.e., a string) of terminals as the repeated application of such functions. At every time step $t$, let us denote the first function mapping each non-terminal to a set of production rules as $k = f(v^t ; \theta_1)$, and the second function mapping each rule to a non-terminal/terminal pair as $(v^{t+1}, w^t) = g(p_i; \theta_2)$ where $v \in V$, and $w \in \Sigma$.
$k$ is a latent vector describing the production rule activations corresponding to $v^t$.

\begin{figure}
    \centering
    \includegraphics[width=\linewidth]{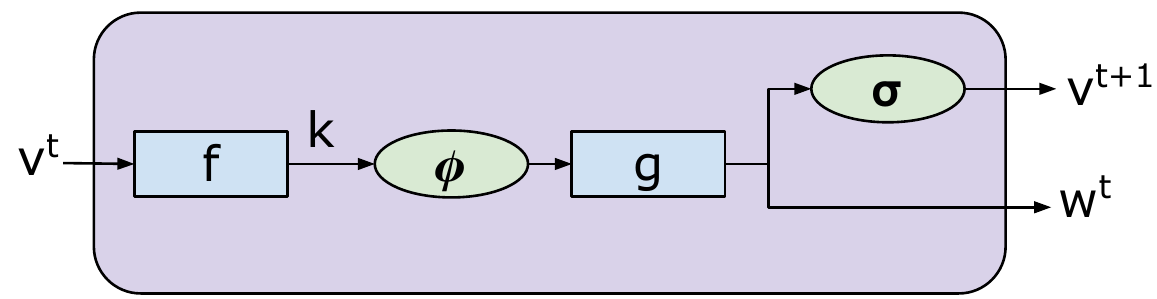}
    \caption{Illustration of the connection between functions in the grammar model. $\phi$ is the gumbel-softmax function and $\sigma$ is the softmax function.}
    \label{fig:grammar-process}
\end{figure}

In its simplest form, we can make our grammar rely only on one production rule by applying the softmax function ($\sigma$) to the activation vector $k$: $p_i = \sigma(k)$. This formulation makes $p_i$ a (soft) one-hot indicator vector selecting the $i$-th production rule.
Our sequence generation then becomes:
\begin{equation}
\big(v^{t+1}, w^t\big) = g\Big(\sigma\big(f(v^t; \theta_1)\big) ; \theta_2 \Big).
\label{eq:single-rule}
\end{equation}
We represent each $v \in V$ as a $N$-dimensional soft one-hot vector where $N$ is the number of non-terminals. In the actual implementation, this is constrained by having a softmax function as a part of $g_{\theta_2}$ to produce $v^{t+1}$. Each $w \in \Sigma$ is a $T$-dimensional representation we learn to generate, where $T$ is the dimensionality of the sequential representation at every time step. This process is shown in Fig. \ref{fig:grammar-process}.

We further extend Eq.~\ref{eq:single-rule} to make the grammar consider multiple production rules in a randomized fashion during its learning and generation. More specifically, we use the Gumbel-Softmax trick  \cite{jang2016categorical,maddison2016concrete} to replace the softmax in Eq.~\ref{eq:single-rule}. Treating the activation vector $k$ as a distribution over production rules, the Gumbel-Softmax ($\phi$) allows sampling of different production rules:
\begin{equation}
\big(v^{t+1}, w^t\big) = g\Big(\phi\big(f(v^t ; \theta_1)\big) ; \theta_2 \Big).
\label{eq:multi-rule}
\end{equation}
In our case, this means that we are learning the grammar production rules which could be selected/sampled differently even for the same non-terminal (i.e., $v^t$) while still maintaining a differentiable process (Fig. \ref{fig:grammar-branching}).

The idea behind our grammar formulation is to allow direct training of the parameters governing generation of the terminals (e.g., video representations in our case), while representing the process in terms of explicit (differentiable) production rules. This is in contrast to traditional work that attempted to extract grammar from already-trained standard RNNs \cite{gers2001lstm} or more recent neural parsing works using 
discrete operators \cite{dyer2016recurrent} and memory-based RNNs \cite{graves2014neural}. Our formulation also adds interpretability to our temporal models learned from data streams, as we confirm more in the following subsections.

\paragraph{Detailed implementation of production rule functions:}

Although any other differentiable functions could be used for modeling our functions $f$ and $g$, we use matrix operations to implement them. Given a matrix of production rules, $W$, a $N\times (R \cdot N)$ matrix, where $R$ is the maximum number of production rules per non-terminal, we obtain the activation vector $k$ with size $R \cdot N$ as:
\begin{equation}
    k = f(v) = v W
\end{equation}
We constrain $W$ so that its each column is a vector with only one non-zero element (i.e., each production rule may originate from only one non-terminal). In the actual implementation, $W$ is obtained by modeling it as a $N\times R$ matrix and then inflating it with zeros to have the form of a block diagonal matrix of size $N\times (R \cdot N)$ with the block size $1\times R$.

Similarly, the function $g$ mapping each production rule to the next non-terminal and corresponding terminal is implemented using a $(R \cdot N) \times N$ matrix $H_1$, and a $(R \cdot N) \times T$ matrix $H_2$:
\begin{equation}
    (v^{t+1}, w^t) = g(v^t) = (\sigma(F H_1), F H_2)
\end{equation}
where $F = \phi(f(v^t))$. With this implementation, learning the grammar production rules is done by learning the matrices $W$, $H_1$, and $H_2$ directly. Fig.~\ref{fig:vis-toy-grammar} describes an example.

\subsection{Learning}

We train our grammar model to minimize the following binary cross entropy loss:
\begin{equation}
    \mathcal{L} = \sum_{t,c} z^t_{c}\log(w^t_{c}) + (1-z^t_{c})\log(1-w^t_{c})
\end{equation}
where $z^t$ is the ground truth label vector at time $t$ with dimensionality $|c|$ and $w^t$ is the output of the grammar model (terminal). In the case where the grammar is used to predict discrete class labels, $z^t$ becomes a one-hot vector. Training of our functions $f$ and $g$ (or matrices $W$, $H_1$, and $H_2$) can be done with a straight forward backpropagation for the simple production rule case of Eq.~\ref{eq:single-rule}, as it becomes a deterministic function per non-terminal at each time step. Backpropagating through the entire sequential application of our functions also allow learning of the starting non-terminal representation $S = v^0$.

\paragraph{Learning multiple production rules:}

In general, our function $f$ maps a non-terminal to a `set' of production rules where different rules could be equally valid. This means that we are required to train the model by generating many sequences, by taking $b$ rules at each step ($b$ is the branching factor).

\begin{figure}
    \centering
    \includegraphics[width=\linewidth]{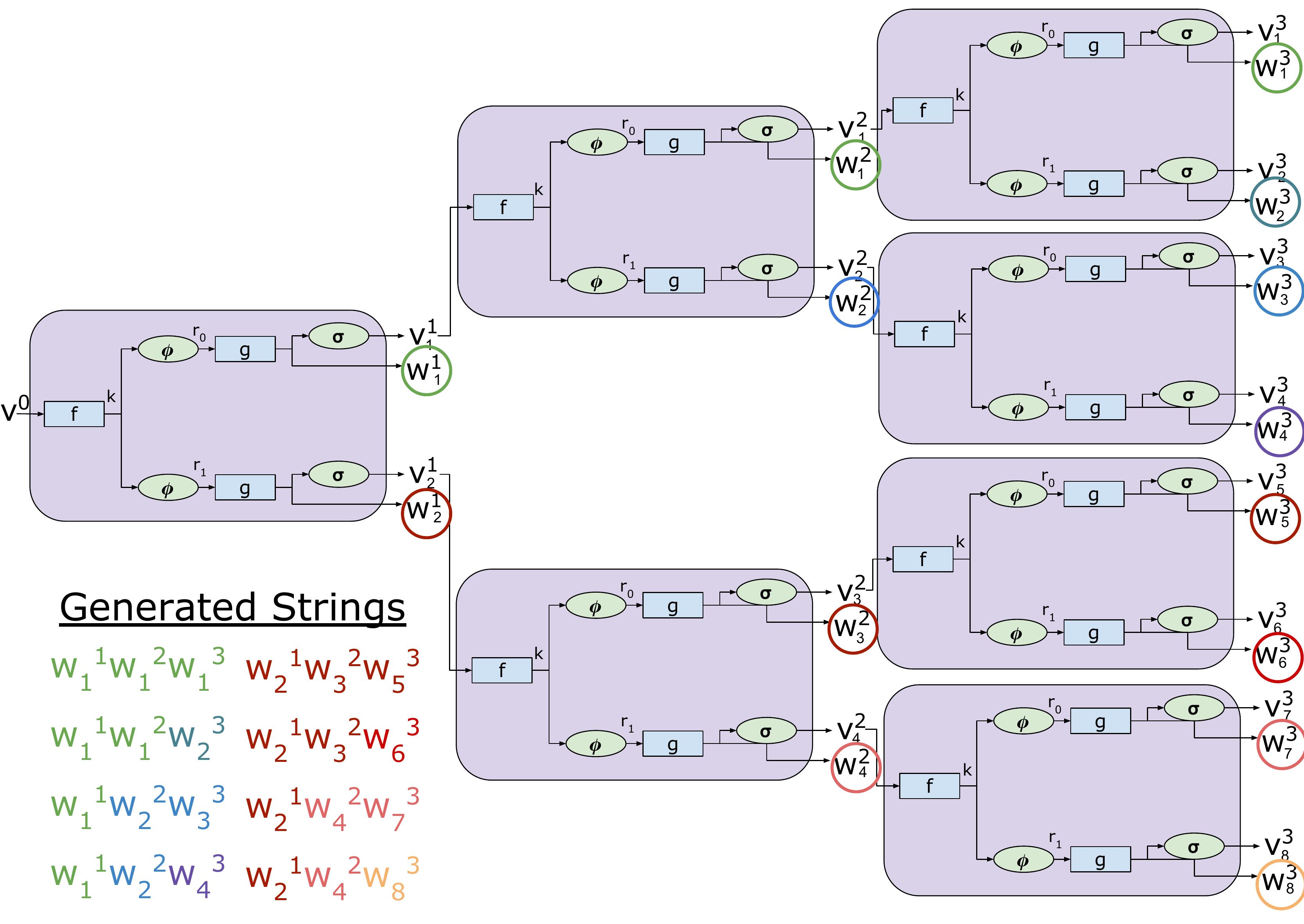}
    \caption{Visualization of training the grammar with branching. The output ($r_i$) of the Gumbel-Softmax ($\phi$) is different for each branch, producing different strings.}
    \label{fig:grammar-branching}
\end{figure}



We enumerate through multiple productions rules by randomizing the production rule selection by using the Gumbel-Softmax trick \cite{jang2016categorical,maddison2016concrete} as suggested in the above subsection. This allows for weighted random selection of the rules based on the learned rule probabilities. In order the train our grammar model with the Gumbel-Softmax, we maintain multiple different `branches' of non-terminal selections and terminal generations, and measure the loss by considering all of them. Algo.~\ref{alg:grammar-training} and Fig. \ref{fig:grammar-branching} illustrate the training and branching process. 
When generating many branches, we compute the loss for each generated sequence, then take the minimum loss over the $b$ branches, effectively choosing the branch that generated the most similar string:
\begin{equation}
    \mathcal{L} = \min_b \sum_{t,c} z_{t,c}\log (w^t_{b,c}) + (1-z_{t,c})\log(1-w^t_{b,c})
\end{equation}
where $w^t_{b,c}$ is the output of the grammar model (terminals) at time $t$ for class $c$ and branch $b$. Branches are pruned to make the process computationally tractable, limiting the total number of branches we maintain.

\begin{algorithm}[tb]
\begin{algorithmic}
   \STATE Input: sequence $s$
   \STATE Set initial nonterminal $v^0$
   \FOR{$t=0$ {\bfseries to} $T$}
   \FOR{$c=0$ {\bfseries to} current total branches}
   \STATE Get rules for current nonterminal: $k = f(v^t_c)$
   \FOR{$b=0$ {\bfseries to} Number of branches}
   \STATE Randomly select a rule: $p = \phi(k)$
   \STATE Get next non-terminal and terminal 
   \STATE $(v^{t+1}_b, w^t_b) = g_{\theta_2}(p)$ 
   \ENDFOR
   \ENDFOR
   \ENDFOR
   \STATE $loss = \min_b \mathcal{L}(s, w_b)$, min over all branches
\end{algorithmic}
 \caption{The training of the grammar, with multiple branches}
    \label{alg:grammar-training}
\end{algorithm}

\subsection{Interpretability}

As our model is constrained to use a finite set of non-terminals, terminals and production rules, it allows for easy interpretability of the learned grammar structure. We can conceptually convert the learned production rule matrices $W$, $H_1$, and $H_2$ into a discrete set of symbolic production rules by associating symbols with the learned terminal (and non-terminal) representations. The matrix $W$ describe the left-hand side non-terminal of the production rule following the regular grammar (e.g., $\textbf{A} \rightarrow aB$), the matrix $H_2$ describes the terminal of the production rule (e.g., $A \rightarrow \textbf{a}B$), and the matrix $H_1$ corresponds to the right-hand side non-terminal of the rule (e.g., $A \rightarrow a\textbf{B}$). Element values of the matrix $W$ in particular suggests the probability associated with the production rule (i.e., it governs the probably of the corresponding production rule being randomly selected with Gumbel-Softmax). Fig. \ref{fig:vis-toy-grammar} shows how we can construct a grammar from the learned matrices.

Figs.~\ref{fig:charades-examples},~\ref{fig:baseball_rule} illustrate examples of such interpreted grammar, learned from raw datasets. This was done by associating symbols with $w^t$ and $v^t$.
We note that the interpretability~\cite{DoshiKim2017Interpretability}  of the model is inherent in our method and comes as an effect of the grammar representation used. 
More specifically, our method is able to embed grammar-like rules into differentiable learning, while still preserving the interepretability of it.


\subsection{Application to video datasets}

Videos contain complex continuous data from with rich visual time-correlated features. We here describe how to apply the grammar model to videos.


The video is first processed by a backbone model which extracts feature representations of chunks of video sequence (e.g., we use the popular I3D model \cite{carreira2017quo}).
The initial non-terminal is learned based on the video representation. We learn a function $\psi$ that maps from the video representation to the initial non-terminal: $S = v_0 = \psi(q)$, where $q$ is the output of the video CNN. We then train the grammar model as above, where the ground truth is the sequence of one-hot vector based on the activity labels in the video.

During inference (which is about predicting frame-level activity labels), we generate a sequence by selecting the rule that best matches the CNN predicted classes. We then multiply the predictions from the grammar with the predictions from the CNN. To predict future actions, we generate a sequence following the most likely production rules.


\section{Experiments}
\begin{figure*}
    \centering
    \includegraphics[width=\linewidth]{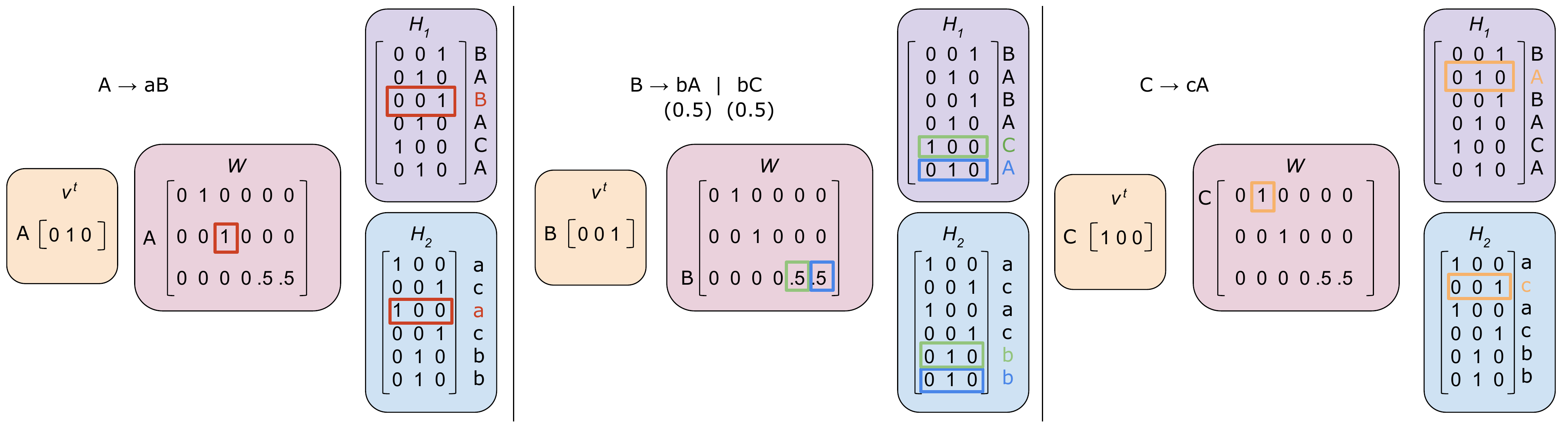}
    \caption{Visualization of the learned toy grammar and how we can construct the grammar from the learned matrices. The non-terminal $v^t$ gives a soft-index into the rule matrix $W$, which gives probabilities over the rules. The rules give a soft-index into the non-terminal matrix ($H_1$) and terminal matrix ($H_2$).}
    \label{fig:vis-toy-grammar}
    \vspace{-3mm}
\end{figure*}

\subsection{Toy Example}
\label{subsec:toy}
We first confirm that our model is able to learn the rules of a simple, hand-crafted grammar and show how we can easily interpret the learned model. Given the grammar:
\begin{align*} 
A&\rightarrow aB\\
B&\rightarrow bC ~|~ bA\\
C&\rightarrow cA
\end{align*}
We train a model with 3 terminal symbols ($a$, $b$, and $c$), 3 non-terminal symbols ($A$, $B$ and $C$), and 2 production rules per non-terminal. We can then examine the learned grammar structure, shown in Fig. \ref{fig:vis-toy-grammar}. We observe that the learned starting non-terminal corresponds to `A', and by following the learned rules, we end up with `aB'. From non-terminal `B', the learned rules go to `bA' or `bC' with 50\% probability. From non-terminal `C', the learned rules go to `cA'. This confirms that the model is able to learn grammar rules and can easily be interpreted.


The approach is general and applies also to any streaming data.

\subsection{Activity Detection Experiments}

We further confirm that our method works on real-world, challenging activity detection datasets: MLB-YouTube \cite{mlbyoutube2018}, Charades \cite{sigurdsson2016hollywood}, and MultiTHUMOS \cite{yeung2015every}. These datasets are evaluated by per-frame mAP. We also compare on 50 Salads \cite{50salads} measuring accuracy. The datasets are described as follows:

\textbf{MultiTHUMOS:} The MultiTHUMOS dataset~\cite{yeung2015every} is a large scale video analysis dataset which has frame-level annotations for activity recognition. It is a challenging dataset and supports dense multi-class annotations (i.e. per frame), which are also used here for both prediction and ground truth. It contains 400 videos or about 30 hours of video and 65 action classes. 


\textbf{Charades:} The Charades dataset~\cite{sigurdsson2016hollywood} is a challenging dataset with unstructured activities in videos. The videos are everyday activities in a home environment. It contains 9858 videos and spans 157 classes. 

\textbf{MLB-YouTube:} The MLB-YouTube dataset~\cite{mlbyoutube2018} is a challenging video activity recognition dataset collected from live TV broadcast baseball games, Fig. \ref{fig:baseball-example}, 
. It further offers the challenge of fine-grained activity recognition as all potential activities are encountered in the same context and environment, unlike many other datasets which feature more diverse activities which may also use context for recognition. It has 4290 videos in 42 hours of video. Additionally, baseball games follow a rigid structure, making it ideal to evaluate the learned grammar. 

\textbf{50 Salads:} This dataset contains 50 videos of making salads, e.g., cutting vegetables and mixing them.

\textbf{Implementation Details} We implemented our models in PyTorch. The learning rate was set to 0.1, decayed every 50 epochs by 10, and the models were trained for 400 epochs. We pruned the number of branches to 2048 by random selection.
The number of grammar parameters vary by dataset driven by the number of classes, 
MLB has 8 terminals (for 8 classes), 5 rules per non-terminal, 8 non-terminals.
Charades - 157 terminals, 10 rules per non-terminal, 1000 non-terminals. The LSTM has 1000 hidden units for all.

\subsection{Results on MLB-Youtube}

Table~\ref{tab:mlb-youtube} shows the results of the proposed algorithm on the MLB-Youtube dataset, compared to all state-of-the-art algorithms including RNNs such as LSTMs.
We evaluated the methods in two different settings: 1) learning grammar on top of features learned from I3D and 2) on top of a recently proposed super-events method. The result clearly shows that our differentiable grammar learning is able to better capture temporal/sequential information in videos. We also compare to LSTMs and NTMs using both CNN features (e.g., I3D) as input and using the predicted class probabilities as input, as that is more comparable to our grammar model. We find that the use of class probabilities slightly degrades performance for LSTMs and NTMs.

\begin{table}
\centering
\caption{Detection results on the MLB-YouTube dataset (mAP).}
\label{tab:mlb-youtube}
\begin{tabular}{|l|c|}
\toprule
\hline
Model                            & mAP \\
\midrule
\hline
Random                           & 13.4 \\
\midrule
I3D                      & 34.2 \\
I3D + LSTM               & 39.4 \\
I3D + NTM (Graves et al. 2014)               & 36.8 \\
I3D class prob + LSTM    & 37.4 \\
I3D class prob + NTM  & 36.8 \\
I3D with Grammar (ours)       & \textbf{43.4} \\
\midrule
I3D + super-events \cite{piergiovanni2018super}      & 39.1 \\
I3D + super-events with Grammar (ours)       & \textbf{44.2} \\
\hline
\bottomrule
\end{tabular}
\vspace{-5mm}
\end{table}

\subsection{Results on MultiTHUMOS}

Table~\ref{tab:main_thumos} shows results comparing two common methods with and without the proposed grammar. We also test both settings as above and compare to the state-of-the-art.
In both settings we can see that use of the learned grammar outperforms previously known methods.

\begin{table} 
  \centering
  \caption{Detection results on the MultiTHUMOS dataset (mAP).} 
    \label{tab:main_thumos}
  \begin{tabular}{|l|c|}
  \toprule
  \hline
  Method  & mAP\\
  \hline
  \midrule 
  Two-stream~\cite{yeung2015every}    & 27.6\\
  Two-stream + LSTM~\cite{yeung2015every}          & 28.1\\
  Multi-LSTM~\cite{yeung2015every}    & 29.6\\
  Predictive-corrective (Dave et al. 2017) & 29.7\\
  \midrule
  I3D baseline                         & 29.7 \\
  I3D + LSTM                           & 29.9 \\
  I3D + NTM (Graves et al. 2014)               & 29.8 \\
  I3D class prob + LSTM    & 29.8 \\
  I3D class prob + NTM     & 29.7 \\
   I3D with Grammar (ours)  &\textbf{32.3} \\
   \midrule
   I3D + super-events~\cite{piergiovanni2018super}            & 36.4\\
   I3D + super-events with Grammar (ours)   & \textbf{37.7} \\
   \midrule
   I3D + TGM~\cite{tgm} & 46.4 \\
   I3D + TGM with Grammar (ours) & \textbf{48.2} \\
   \hline
  \bottomrule
  \end{tabular}
  \vspace{-4mm}
\end{table}

\subsection{Results on Charades}

Table~\ref{tab:main_charades} has results comparing the proposed grammar to other prior techniques on the Charades dataset (v1\_localize setting). As seen, this dataset is quite challenging since recently its detection accuracy was below 10 percent mAP. 
Our results here too outperform the state-of-the-art, increasing the accuracy on this dataset. We note that there are consistent improvements in both settings, similar to the results on MultiTHUMOS and MLB-YouTube. In particular, the differentiable grammar learning outperformed previous RNNs including LSTMs and NTMs.

\begin{table} 
  \centering
 \caption{Detection results on the Charades dataset (Charades\_v1\_localize setting).} 
\label{tab:main_charades}
  \begin{tabular}{|l|c|}
  \toprule
  \hline
  Method  & mAP\\
  \hline
  \midrule
  Predictive-corrective (Dave et al. 2017) & 8.9 \\
  Two-stream~\cite{sigurdsson2016asynchronous}& 8.94 \\
  Two-stream+LSTM~\cite{sigurdsson2016asynchronous}& 9.6 \\
  R-C3D~\cite{xu2017r}                         &  12.7 \\
  Sigurdsson et al.~\cite{sigurdsson2016asynchronous} & 12.8 \\
  \midrule
  I3D baseline                                 & 17.2 \\
  I3D + LSTM                                   & 18.1 \\
  I3D + NTM                & 17.5 \\
  I3D class prob + LSTM    & 17.6 \\
  I3D class prob + NTM    & 17.4 \\


   I3D with Grammar (ours)  &\textbf{18.5} \\
   \midrule
   I3D + super-events~\cite{piergiovanni2018super} & 19.4 \\
   I3D + super-events with Grammar (ours)   &\textbf{20.3} \\
   \midrule
   I3D + TGM~\cite{piergiovanni2018super}            & 22.3 \\
   I3D + TGM with Grammar (ours) & \textbf{22.9} \\
\hline
  \bottomrule
  \end{tabular}
\end{table}

\subsection{Future Prediction}

As our grammar model is generative, we can apply it to predict the future, unseen activities. Future prediction is important, especially for autonomous systems (e.g., robots) as they need to anticipate potential future activities to respond to. Once the grammar is learned, future sequences containing unseen activities can be generated by selecting the most probable production rule at every (future) time step. 

For this experiment we consider predicting at short-term horizons (in the next 2 seconds), mid-term horizons (10 seconds), and more longer-term horizons (20 seconds). We compare to baselines such as random guessing, repeatedly predicting the last seen frame, and an LSTM approach (with I3D features) which is commonly used for future frame forecasting. We evaluate these methods using per-frame mAP.

Table~\ref{tab:fut_thumos} shows the results for future prediction on the MultiTHUMOS dataset. We confirm the grammar generates more accurate futures.
We note that 10-20 seconds in the future is a very challenging setting to try to predict especially in the context of multi-label datasets.

Table~\ref{tab:fut_charades} shows the results for future prediction for the Charades dataset. Here too, we can see the proposed grammar approach is more accurate at future frame prediction.

Following the setting in \cite{ke2019future}, we evaluate the future prediction task on the 50 Salads dataset \cite{50salads}. The results are shown in Table \ref{tab:salad}. The results confirm that our approach allows better prediction.

Previous work on video forecasting, e.g., \cite{yeung2015every} did not address longer horizons.
Other work on early detection is also related ~\cite{ryoo11,ma2016learning,want2018early}, but they only focused on early detection of an ongoing event rather than forecasting multiple longer-term future events.

\begin{table} 
  \centering
  \caption{Future prediction on the MultiTHUMOS dataset for various time horizons.} 
  \label{tab:fut_thumos}
  \begin{tabular}{|l|c|c|c|}
  \hline
  Method   &\cellcolor{col1} 2 sec &\cellcolor{col2} 10 sec  & \cellcolor{col3} 20 sec\\
  \hline 
   Random    & 2.6 & 2.6 & 2.6 \\
   Last frame & 16.0 & 12.7 & 8.7 \\
   I3D + LSTM   &15.7 &6.8 &4.1 \\
   I3D + Grammar (ours)   &\textbf{18.6} &\textbf{12.8} &\textbf{10.5} \\
  \hline
  \hline
  \end{tabular}
  \vspace{-4mm}
\end{table}


\begin{table} 
  \centering
  \caption{Future prediction on the Charades dataset for various time horizons.}
  \label{tab:fut_charades}
  \begin{tabular}{|l|c|c|c|}
  \hline
  Method   &\cellcolor{col1} 2 sec &\cellcolor{col2} 10 sec  & \cellcolor{col3} 20 sec\\
  \hline 
   Random    & 2.4 & 2.4 & 2.4 \\
   Last frame & 13.8 & 11.2 & \textbf{8.6} \\
   I3D + LSTM   &12.7 &10.8 &7.0 \\
   I3D + Grammar (ours)   &\textbf{14.8} &\textbf{11.2} & 8.5 \\
  \hline
  \hline
  \end{tabular}
  \vspace{-3mm}
\end{table}

\begin{table}[]
    \centering
    \small
    \caption{Results on 50 Salads without ground-truth observations (setting in Table 2 of \cite{ke2019future}).}
    \begin{tabular}{l@{\hspace{4pt}}c@{\hspace{4pt}}c@{\hspace{4pt}}c@{\hspace{4pt}}c@{\hspace{4pt}}c@{\hspace{4pt}}c@{\hspace{4pt}}c@{\hspace{4pt}}c@{\hspace{4pt}}}
    \toprule
      Observation & \multicolumn{4}{c}{20\%} & \multicolumn{4}{c}{30\%} \\
      \cmidrule(lr){2-5}\cmidrule(lr){6-9}
      Prediction & 10\% & 20\% & 30\% & 50\% & 10\% & 20\% & 30\% & 50\% \\
      \midrule
      RNN               & 30.1 & 25.4 & 18.7 & 13.5 & 30.8 & 17.2 & 14.8 & 9.8 \\
      CNN               & 21.2 & 19.0 & 16.0 & 9.9 & 29.1 & 20.1 & 17.5 & 10.9 \\
      TCA (Ke et al.) & 32.5 & 27.6 & 21.3 & 16.0 & 35.1 & 27.1 & 22.1 & 15.6\\
\hline
    Grammar       & \textbf{39.2} & \textbf{32.1} & \textbf{24.8} & \textbf{19.3} & \textbf{38.4} & \textbf{29.5} & \textbf{25.5} & \textbf{18.5} \\
      \bottomrule
    \end{tabular}
    \label{tab:salad}
\end{table}

\begin{figure}
    \centering
    \includegraphics[width=\linewidth]{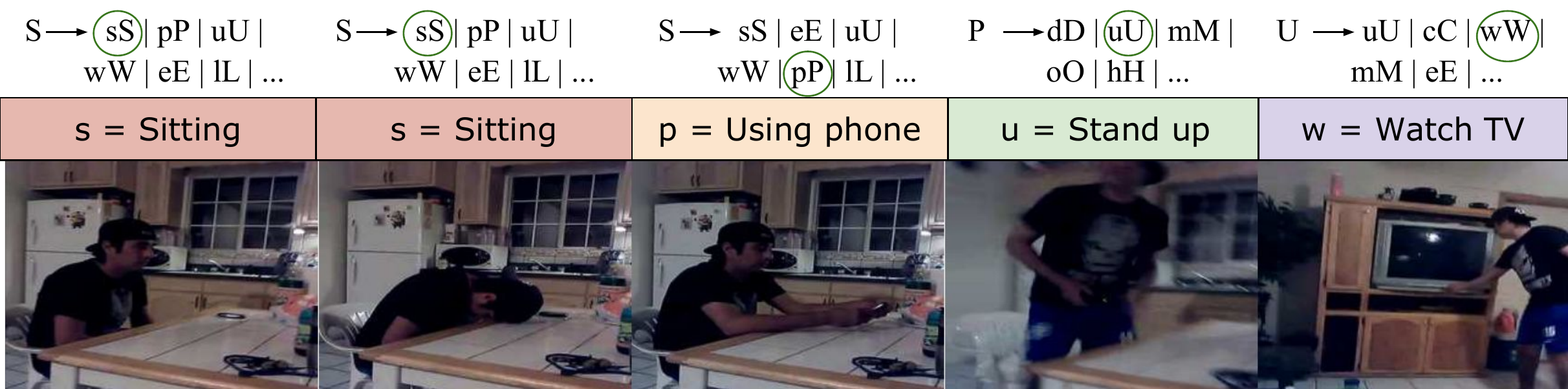}
    \caption{Examples of a sequence and grammar structure learned from the Charades dataset. Other possible activities include e=eating, h=holding cup, m=making sandwich, l=using laptop, etc.}
    \label{fig:charades-examples}
\end{figure} 
\vspace{-0.2cm}
  

\subsection{Visualization of Learned Grammars}
In Fig. \ref{fig:vis-toy-grammar}, we illustrate how we convert from the learned matrices to the grammar and production rules. From the training data, we know the mapping from terminal symbol to label. We can then examine the rule matrix, $W$ and the non-terminals, $H_1$ to construct the rules.

Fig.~\ref{fig:charades-examples} visualizes example learned grammar rules for the Charades dataset. We can see learned grammar rules corresponding to natural sequences of events.

\begin{figure}
    \centering
    \includegraphics[width=0.77\linewidth]{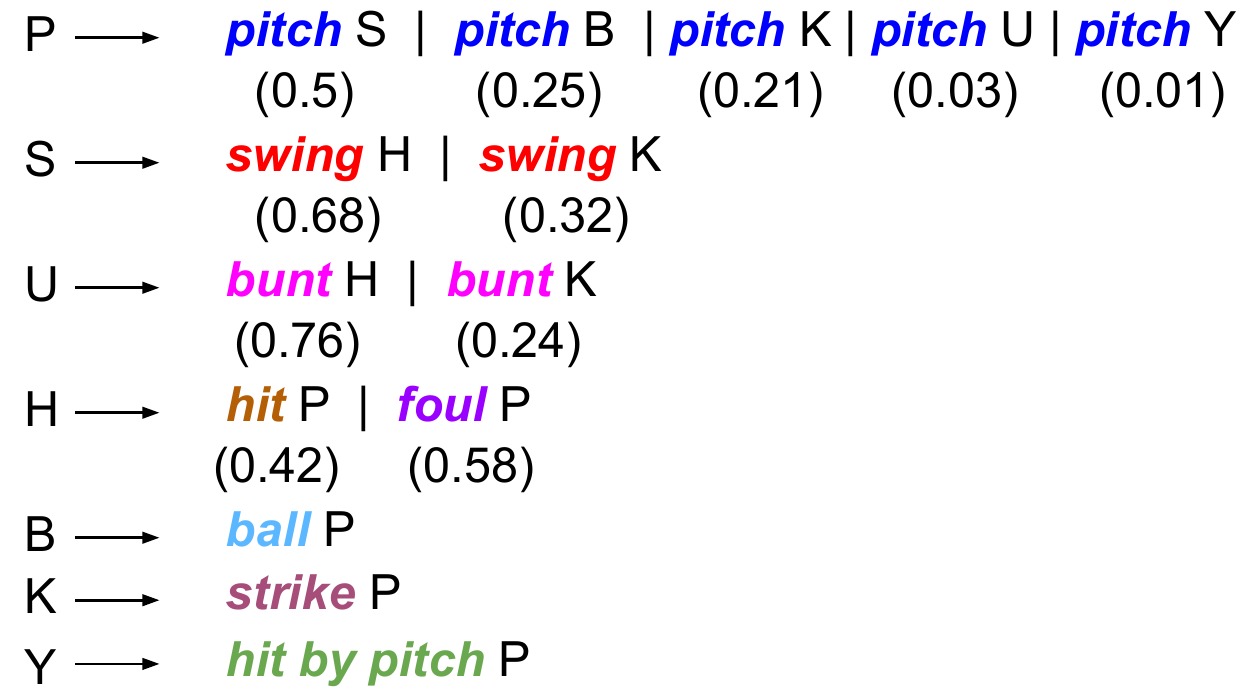}
    \caption{The learned grammar from the MLB-Youtube dataset. For non-terminals with multiple rules, the learned probabilities are in parenthesis.}
    \label{fig:baseball_rule}
    \vspace{-4mm}
\end{figure}

We also visualize the learned grammar for the MLB-YouTube dataset, in which, since it is smaller, we can extract all the learned rules. Fig.~\ref{fig:baseball_rule} is the conceptual visualization of the learned regular grammar. Interestingly the typical baseball sequences are learned.  
More specifically, in Fig. \ref{fig:baseball_rule}, we see that that the learned grammar matches the structure in a baseball game and the probabilities are similar to the observed data, confirming that our model is able to learn the correct rule structure. For example, an activity starts with a pitch which can be followed by a swing, bunt or a hit. After a hit, foul, or strike, another pitch follows. The learned grammar is illustrated with probabilities for each rule in parenthesis. In the appendix, we 
illustrate the actual learned matrices corresponding to one of the production rules.


\vspace{-0.3cm}




\section{Conclusion}

In conclusion, we presented a differentiable model for learning formal grammars
for the purposes of video understanding. The learned structures are interpretable which is important for understanding the behavior of the model and the decisions made.
The proposed method outperforms all prior state-of-the-art techniques on 
several challenging benchmarks.
Furthermore, it can predict future events with higher accuracy, which is  necessary for anticipation and reaction to future actions.
In the future we plan to apply it to even longer horizon data streams. Further, we aim to enable application of our differentiable grammar learning to higher-dimensional representations, learning them jointly with image and video CNNs in an end-to-end fashion.



{\small
\bibliography{egbib}
\bibliographystyle{aaai}
}

\end{document}


\maketitle

\section{Details on Grammar Process}
\begin{figure}
    \centering
    \includegraphics[width=0.8\linewidth]{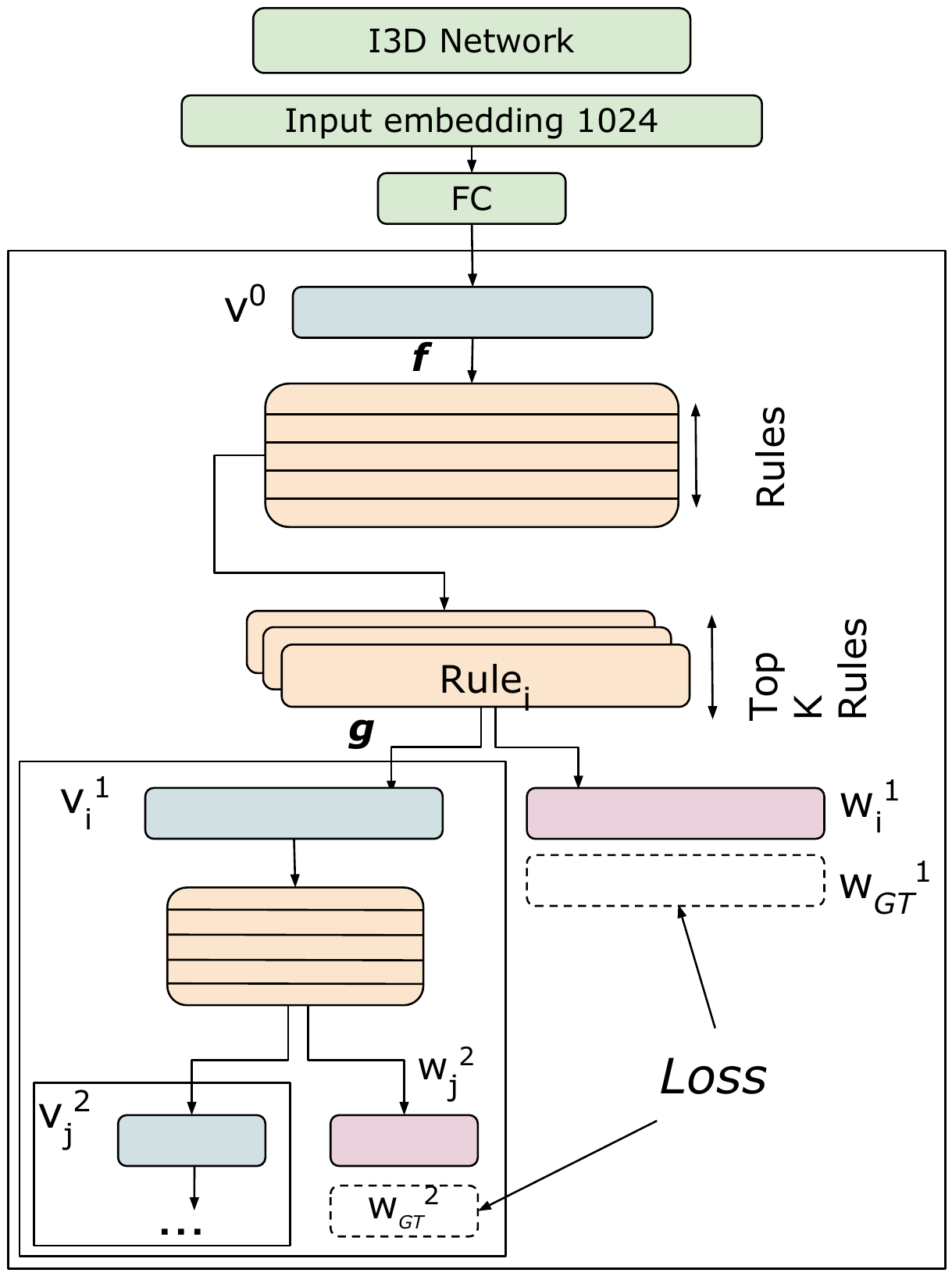}
    \caption{Learning of the video grammar model. At each step, the model maps a non-terminal to production rules each generating a terminal and a non-terminal.}
    \label{fig:grammar-main}
\end{figure}

These form a straight forward (recursive) generative model, which starts from the starting non-terminal $S = v^0$ and iteratively generates a terminal at every time step. Figure~\ref{fig:grammar-main} visualizes the schematic of the process: the system internally keeps states of possible rules and non-terminals and learns their relationships with terminals according to data, starting from a feature representation of the video, non-terminal `states' $v$ are learned, each one of which learns to generate a number of rules. Each rule is then expanded into a set of non-terminal and terminal, where the terminal can now be compared to the actually observed data (terminal value). This process continues recursively.

\subsection{Visualization of Learned Grammars}

\begin{figure*}
    \centering
    \includegraphics[width=\linewidth]{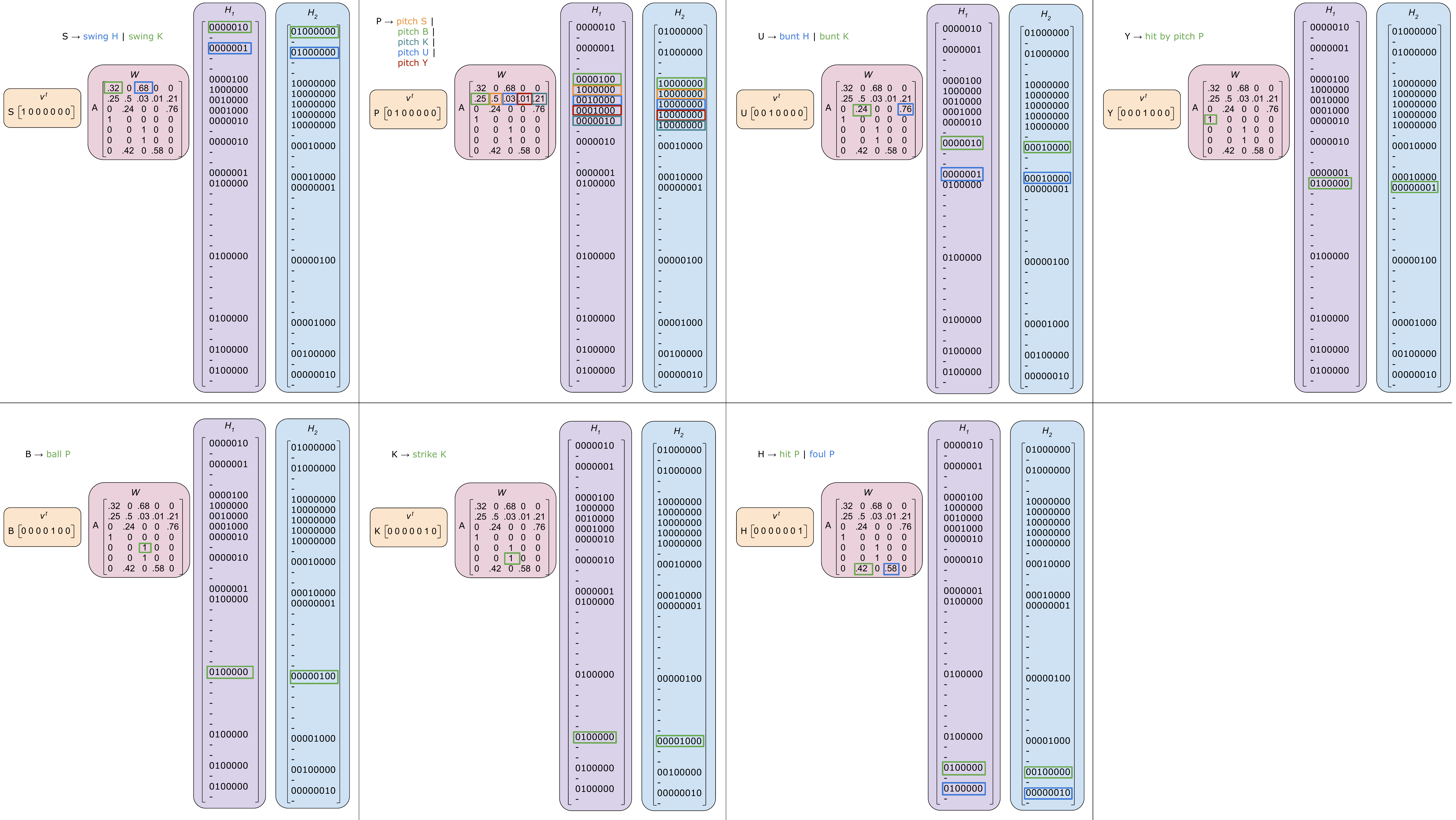}
    \caption{Visualization of the learned grammar for MLB-YouTube videos. Note, for simplicity, we only visualize the applicable rules and are omitting zeros in the block-diagonal matrix $W$. `-' is used for terminals and non-terminals that are never used.}
    \label{fig:vis-mlb-youtube}
\end{figure*}

We also visualize the additional examples of the learned grammar.  In Figure \ref{fig:vis-mlb-youtube}, we illustrate how all the learned rules are inferred from the learned matrices.

\begin{figure}
    \centering
    \includegraphics[width=\linewidth]{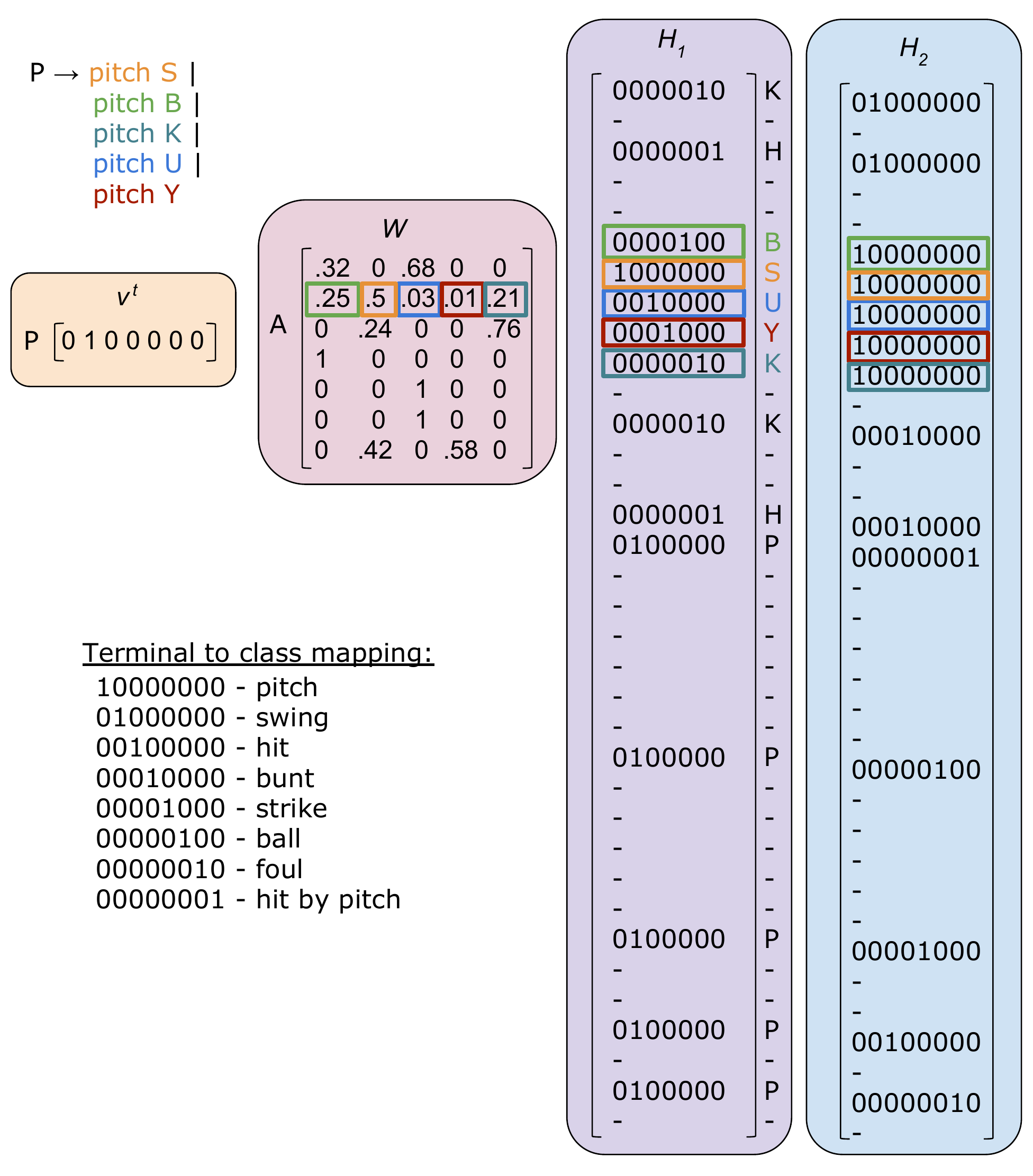}
    \caption{Visualization of one learned non-terminal rule pair. For simplicity, we only visualize the rules that are applicable for each non-terminal. `-' denotes terminals and non-terminals that are never used.}
    \label{fig:one-baseball-rule}
\end{figure}

Figure \ref{fig:one-baseball-rule} illustrates the actual learned matrices corresponding to one of the production rules.

\begin{figure*}
    \centering
    \includegraphics[width=\linewidth]{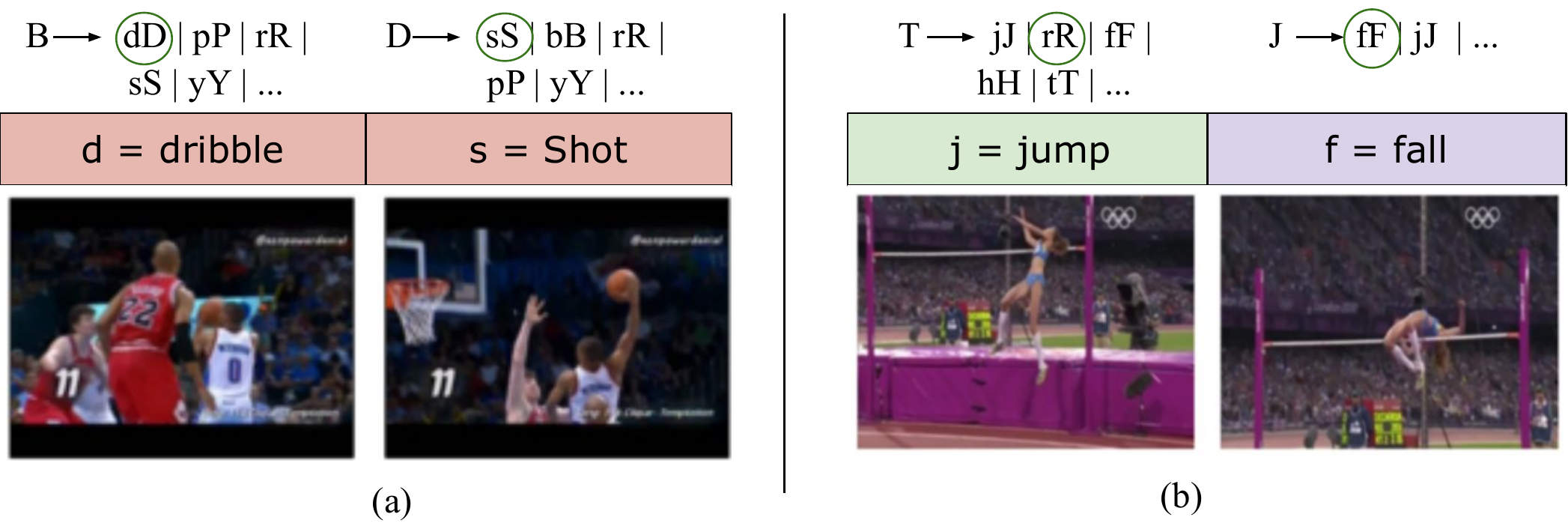}
    \caption{Example of grammar sequences from MultiTHUMOS videos. B and T are the starting non-terminals (B for basketball videos and T for track and field videos). The other actions are d for dunking, p for passing, r for running, s for shooting, y for guarding, j for jumping, f for falling, h for hurdles, and t for throwing.}
    \label{fig:mt-example}
\end{figure*}

Figure~\ref{fig:mt-example} visualizes example learned grammar rules for the MultiTHUMOS dataset. While, this is a big dataset, we can see learned grammar rules corresponding to natural sequences of events.



